\pdfoutput=1

\documentclass[11pt]{article}

\usepackage{EMNLP2022}

\usepackage{times}
\usepackage{latexsym}

\usepackage[T1]{fontenc}

\usepackage[utf8]{inputenc}

\usepackage{microtype}

\usepackage{inconsolata}

\usepackage{url}
\usepackage{amsmath}
\usepackage{amssymb}
\usepackage{booktabs}
\usepackage{graphicx}
\usepackage{varwidth}
\usepackage{tabularx}
\usepackage{algorithm}
\usepackage[noend]{algpseudocode}
\usepackage{caption}
\usepackage{adjustbox}

\usepackage[utf8]{inputenc}

\usepackage[textsize=scriptsize]{todonotes}

\DeclareMathOperator*{\argmax}{argmax}

\title{Natural Language Deduction through Search over Statement Compositions}

\author{Kaj Bostrom \quad\quad
  Zayne Sprague \quad\quad
  Swarat Chaudhuri \quad\quad
  Greg Durrett \\
  Department of Computer Science\\
  The University of Texas at Austin\\
  \texttt{\{kaj,swarat,gdurrett\}@cs.utexas.edu, zaynesprague@utexas.edu}
  }

\date{}

\begin{document}
\maketitle
\begin{abstract}
In settings from fact-checking to question answering, we frequently want to know whether a collection of evidence (premises) entails a hypothesis. Existing methods primarily focus on the end-to-end discriminative version of this task, but less work has treated the \emph{generative} version in which a model searches over the space of statements entailed by the premises to constructively derive the hypothesis. We propose a system for doing this kind of deductive reasoning in natural language by decomposing the task into separate steps coordinated by a search procedure, producing a tree of intermediate conclusions that faithfully reflects the system's reasoning process. Our experiments on the EntailmentBank dataset \cite{dalvi-etal-2021-explaining}  demonstrate that the proposed system can successfully prove true statements while rejecting false ones. Moreover, it produces natural language explanations with a 17\% absolute higher step validity than those produced by an end-to-end T5 model.
\end{abstract}

\section{Introduction}

When we read a passage from a novel, a Wikipedia entry, or any other piece of text, we gather meaning from it beyond what is written on the page. We make inferences based on the text by combining information across multiple statements and by applying our background knowledge. This ability to synthesize meaning and determine the consequences of a set of statements is a significant part of natural language understanding. Humans are able to give step-by-step explanations of the reasoning that they do as part of these processes. However, approaches that involve end-to-end discriminative fine-tuning of pre-trained language models have no such notion of step-by-step deduction; these models are black boxes and do not offer explanations for their predictions. This limitation prevents users from understanding and accommodating models' affordances \cite{hase-bansal-2020-evaluating,BansalEtAl2021}.

\begin{figure}
    \centering
    \includegraphics[width=\columnwidth]{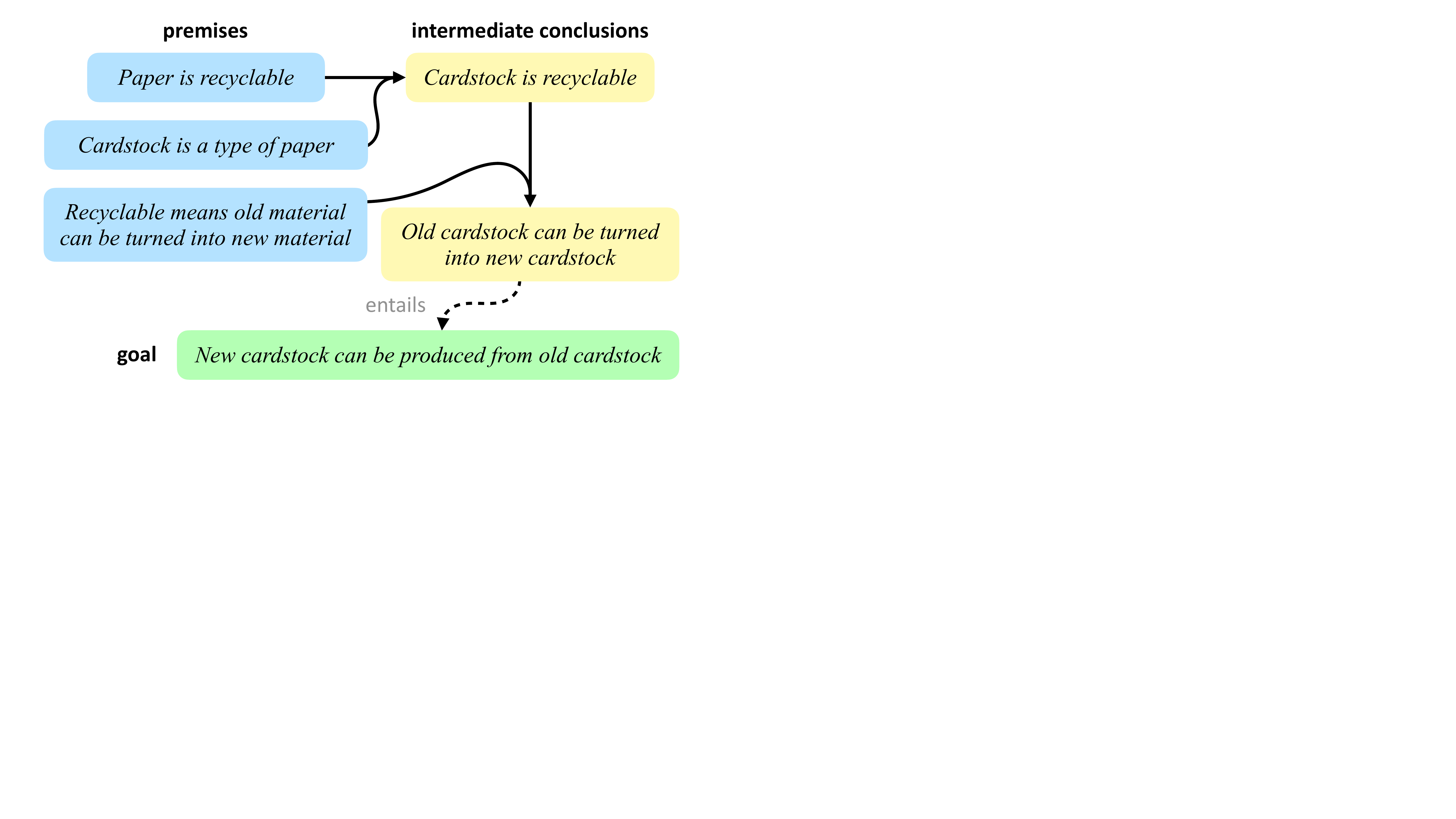}
    \caption{An example of multi-step natural language deduction performed by our system. From input premises, our model generates new statements in a heuristic-guided way to try to prove a given hypothesis.}
    \label{fig:example}
\end{figure}

\begin{figure*}[t!]
    \centering
    \includegraphics[width=\textwidth]{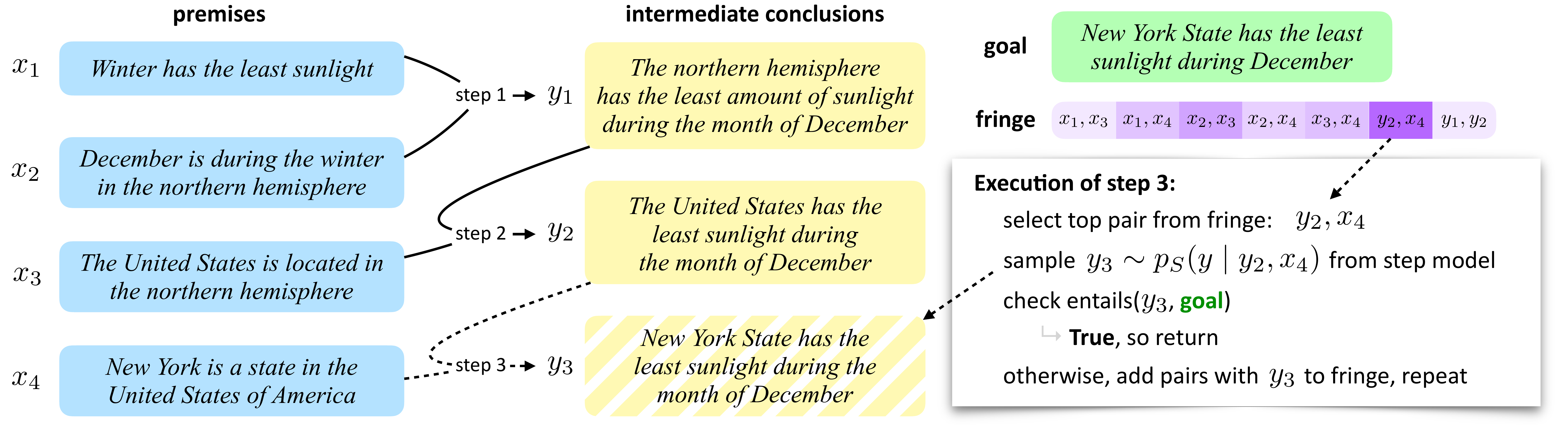}
    \caption{Overview of the problem setting and our approach. The search fringe tracks which statements should be combined next, scored according to the heuristic $h$ (indicated here by purple shading). See Algorithm \ref{alg:spc} for a detailed account of the search procedure.}
    \label{fig:overview}
\end{figure*}

A simple way of representing this kind of reasoning process is through an entailment tree \cite{dalvi-etal-2021-explaining}: a derivation indicating how each intermediate conclusion was composed from its premises, exemplified in Figure \ref{fig:example}.
Generative sequence-to-sequence models can be fine-tuned to carry out this task given example trees \cite{tafjord-etal-2021-proofwriter, dalvi-etal-2021-explaining}. However, we argue in this paper that this conflates the generation of individual reasoning steps with the planning of the overall reasoning process. An end-to-end model is encouraged by its training objective to generate steps that arrive at the goal, but it is not constrained to do so by following a sound structure. As we will show, the outputs of such methods may skip steps or draw unsound conclusions from unrelated premises while claiming a hypothesis is proven. Generating an explanation is not enough; we need explanations to be consistent and faithfully reflect a reasoning process to which the model commits \cite{jacovi-goldberg-2020-towards}.

This paper proposes a system that factors the reasoning process into discrete search over intermediate steps. The core of our system is a step deduction module that generates the direct consequence of composing a pair of statements. This module is used as a primitive in a search procedure over entailed statements, guided by a learned heuristic function. By decoupling the deduction itself from the search over statement combinations, our system's design ensures that each step's conclusion builds on its inputs, avoiding the pitfalls of end-to-end generation.

We evaluate our method on the EntailmentBank dataset \cite{dalvi-etal-2021-explaining}. Thanks to our system's factored design and its ability to capitalize on additional semi-synthetic single step data \cite{bostrom-etal-2021-flexible}, we observe that 82\% of the reasoning steps produced by our approach are sound, a 17\% absolute increase compared to an end-to-end model replicated from prior work.

Our contributions are: (1) A factored, interpretable system for natural language deduction, separating the concerns of generating intermediate conclusions from those of planning entailment tree structures; (2) Exploration of several search heuristics, including a learned goal-oriented heuristic; (3) Comparison to an end-to-end model from prior work \cite{dalvi-etal-2021-explaining} in two settings of varying difficulty.

\section{Problem Description and Motivation}

The general setting we consider is shown in Figure \ref{fig:overview}. We assume we are given a collection of evidence sentences $X = \{x_1\dots x_n\}$\footnote{In the settings we consider in this paper, $n$ does not exceed 25. However, this is not a hard limit imposed by our approach, and for very large premise sets, a retrieval system could be used to prune the premise set to a manageable size.} and a goal statement $g$. We want to construct an \textbf{entailment tree} deriving $g$ from $X$. An entailment tree is a tree of statements with the property that each statement is directly entailed by its children. Thus, if we can produce a tree with root $g$ and leaves $X$, it follows that $g$ is transitively entailed by the premises in $X$.

Crucially, we assume that the intermediate nodes of this tree must be generated and are not present in the input premise set $X$. This condition differs from prior work on question answering with multi-hop reasoning \cite{yang-etal-2018-hotpotqa,chen2021multihop} or models that build a proof structure but do not generate new statements \cite{saha-etal-2020-prover,saha-etal-2021-multiprover}. We therefore require a generative step deduction model $S$ to produce intermediate conclusions given their immediate children, a concept explored by  \citet{tafjord-etal-2021-proofwriter} and \citet{bostrom-etal-2021-flexible}.

$S$ yields a distribution $p_S(y\ |\ x_1 \dots x_m)$ over step conclusions $y$ conditioned on inputs $x_i$. In our approach, we assume that each step has exactly $m=2$ inputs. Some nodes in our evaluation dataset, EntailmentBank, have more than two children, but in preliminary investigation we found it possible to express the reasoning in these steps in a binary branching format. We do not apply this arity constraint to baseline models.

\section{Methods}

Our proposed system consists of a step model $S$, a search procedure involving a heuristic $h$, and an entailment model which judges whether a generated conclusion entails the goal. An overview of the responsibilities and task data required by each module is presented in Table \ref{tab:task_data}.

\begin{table}[]
    \small
    \centering
    \begin{tabular}{c c}
        \toprule
        \textbf{Module} & \begin{tabular}{@{}c@{}} \textbf{Task data format} \\ \textbf{(Datasets used)} \end{tabular} \\
        \midrule
        Step model $S$ & \begin{tabular}{@{}c@{}} Single-step deductions \\ (ParaPattern, EntailmentBank) \end{tabular} \\ \midrule
        Heuristic $h$ & \begin{tabular}{@{}c@{}} Entailment trees* \\ (EntailmentBank) \end{tabular} \\ \midrule
        Goal entailment model & \begin{tabular}{@{}c@{}} Single-sentence NLI \\ (WANLI, EBEntail) \end{tabular} \\
        \bottomrule
    \end{tabular}
    \caption{Each module in our proposed system operates independently, and most components can leverage existing resources without the need for full tree data. *Our best-performing parametric heuristic uses full tree supervision, but nonparametric heuristics are supported.}
    \label{tab:task_data}
\end{table}

\subsection{Step Deduction Model}

Our step models are instances of the T5 pre-trained sequence-to-sequence model \cite{raffel-etal-2020-t5} fine-tuned on a combination of deduction step datasets from prior work \cite{dalvi-etal-2021-explaining,bostrom-etal-2021-flexible}, which we discuss further in Section~\ref{sec:impl_training}. At inference time, we decode from these models using nucleus sampling \cite{Holtzman2020The} with $p=0.9$.

\begin{figure*}
    \centering
    \includegraphics[width=\textwidth]{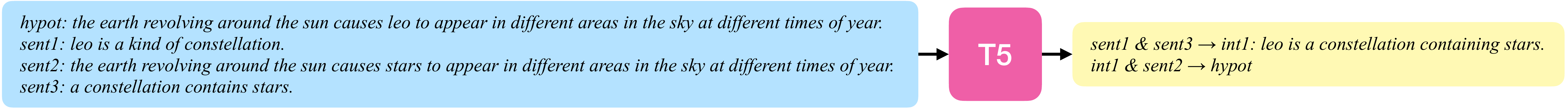}
    \caption{An example of the EntailmentWriter end-to-end system's linearized input and output format. EntailmentWriter takes the goal hypothesis as input, making it possible to hallucinate content based on it, and generation of \emph{$\rightarrow$ hypot} may occur prematurely, before enough evidence is included to truly derive the goal.}
    \label{fig:e2e_format}
\end{figure*}

\subsection{Search}

We define a search procedure over sentence compositions that we call \textsc{SCSearch}, described in Algorithm~\ref{alg:spc}. \textsc{SCSearch} is a best-first search procedure where the search fringe data structure is a max-heap of statement pairs ordered by the heuristic scoring function $h$. \textsc{SCSearch} iteratively combines statement pairs using the step deduction model, adding generated conclusions to the forest of entailed statements.

The heuristic function ${h(\{x_1 \dots x_m\},\ g)\rightarrow \mathbb{R}}$ accepts candidate step inputs $x_i$ and a goal hypothesis $g$ and returns a real-valued score indicating the priority of the potential step in the expansion order.

These priority values reflect multiple factors. We want to prioritize \textbf{compatible} compositions:  combining statements from which we can make a meaningful inference, as opposed to unrelated sentences. We also want to prioritize \textbf{useful} steps: among compatible compositions, we should prefer those that are likely to derive statements that help prove the hypothesis.

\begin{algorithm}[t!]
\small
\caption{}
\begin{algorithmic}
\State {\textbf{procedure} $\textsc{SCSearch}(X = \{x_1\dots x_n\},\ g)$:}
\State $fringe \gets \{\langle x_i, x_j \rangle\ |\ x_i, x_j \in X, i \neq j \}$
\State $visited \gets \{\}$
\State $i \gets 1$
\State $maxSteps \in \mathbb{N}$
\While{$|fringe| > 0 \wedge i \leq maxSteps$}
    \State $inputs \gets \argmax\limits_{\langle x_1, x_2 \rangle \in fringe} h(\{x_1, x_2\}, g)$
    \State $fringe \gets fringe \setminus \{inputs\}$
    \State $\text{sample}\ y_i\ \text{from}\ p_S(y\ |\ inputs)$
    \If{$y_i \notin visited$}
        \State $visited \gets visited\ \cup\ \{y'\}$
        \State $\textbf{yield}\ \langle inputs, y_i \rangle$
        \If{$\text{entails}(y_i, g)$}
            \Return
        \EndIf
        \State $\begin{aligned} fringe \gets fringe & \cup \{ \langle y_i, x_j \rangle | x_j \in X \} \\ & \cup \{ \langle y_i, y_j \rangle | 1 \leq j < i \} \end{aligned} $
        \State $i \gets i + 1$
    
    \EndIf
\EndWhile
\end{algorithmic}
\label{alg:spc}
\end{algorithm}

We consider several potential realizations of the heuristic function $h$:

\paragraph{Breadth-first} Naively, the earliest fringe items are explored first; all compositions of initial premises will be explored before any composition involving an intermediate conclusion.

\paragraph{Overlap} This heuristic scores potential steps according to the number of tokens shared between input sentences. This heuristic is focused on compatibility, as overlap indicates expressions that might be unifiable (e.g., \emph{paper} in the first composition step of Figure~\ref{fig:example}). In the \textbf{Overlap+Goal} version, token overlap with the goal hypothesis is also incorporated into the score.

\paragraph{Repetition} Past work on step deduction models \cite{bostrom-etal-2021-flexible} has identified that these models tend to ``back off'' to copying the input when given incompatible premises.

This heuristic aims to exploit this behavior as a measure of premise compatibility. Potential steps are scored according to $-p_S(x_1 | x_1 ... x_n)$, i.e., the negative likelihood of repeating the first input.

\paragraph{Learned} This heuristic uses an additional pre-trained model fine-tuned to predict whether input statements are part of a gold explanation of the hypothesis or not. We train this model on sets of step inputs drawn from a collection of valid steps augmented with negative samples produced by replacing one input with a random statement.

The \textbf{Learned+Goal} version of this heuristic is also trained with the goal hypothesis in its input, so as to be able to select useful premises and guide search towards the goal. Note that the step model $S$ which generates statements still does not see the goal; the goal only informs which compositions are explored first during the search. See Appendix~\ref{sec:appendix} for training details of our learned heuristic models.

\begin{table*}[t!]
    \centering
    \small
    \begin{tabular}{m{18em} m{18em} c}
        \toprule
        \centering{\textbf{Statement}} & \centering{\textbf{Goal Hypothesis}} & \textbf{Label} \\
        \midrule
        A fish is a kind of scaled animal that uses their scales to defend themselves. & Scales are used for protection by fish. & Entailment\\ \midrule
        Information in an organism's chromosomes causes an inherited characteristic to be passed from parent to offspring by dna.& Children usually resemble parents. & Neutral\\ \midrule
        A rheumatoid arthritis will happen first before other diseases affect the body's tissues. & The immune system becomes disordered first before rheumatoid arthritis occurs. & Neutral \\
        \bottomrule
    \end{tabular}
    \caption{Examples of instances from the \textsc{EBEntail} dataset. The goal hypotheses are from EntailmentBank, and the statements are from the \textsc{SCSearch} algorithm applied to the original EntailmentBank premises. Although the third example contains a contradiction, the `neutral' and `contradiction' labels are merged in \textsc{EBEntail}, as both reflect a failure to entail the goal.}
    \label{tab:goalentailmentdata}
\end{table*}

\subsection{Goal Entailment}
\label{sec:goal_ent}

In order to determine when the search has succeeded, we need a module to judge the entailment relationship between each generated conclusion and the goal. We use a DeBERTa model \cite{he2021deberta} fine-tuned on the WANLI dataset \cite{liu2022wanli} to predict the probability that each derived statement entails the goal. In order to mitigate the domain mismatch between WANLI and the scientific facts that make up EntailmentBank, we also fine-tune our goal entailment model on \textsc{EBEntail}, a set of 300 examples of generated conclusions sampled from \textsc{SCSearch} paired with corresponding goal hypotheses which we manually label for their entailment relationship.

\paragraph{\textsc{EBEntail}} To produce a set of reference judgments for threshold selection and entailment model evaluation, we sampled 150 instances of generated conclusions from \textsc{SCSearch} inference over EntailmentBank examples with their corresponding gold goals. Three annotators labeled the entailment relationship between these generated inferences and the goal. We select a consensus annotation with a majority vote. These examples form an in-domain evaluation set which we call \textsc{EBEntail}. To simultaneously train and evaluate on these judgements, we use 3-fold cross-validation where each cross-validation fold is constructed to contain goals not seen in its respective training fold.

We extend \textsc{EBEntail} with \textsc{EBEntail-Active}, consisting of 150 additional instances with the lowest confidence (highest prediction entropy) following the initial fine-tuning, which are then manually labeled by at least one annotator.

\paragraph{Thresholding} During inference, rather than returning the highest-scoring class, a threshold value $\alpha$ is applied to the predicted probability of the `entailment' class. This threshold allows for better control over trade-off between precision and recall. For our main experiments, we use an entailment score threshold of $\alpha=0.81$ selected via cross-validation on \textsc{EBEntail}.

\section{Experimental Setup}

Our experiments assess whether our \textsc{SCSearch} system, which factors the deduction process into separate step generation and search modules, can do better than end-to-end baselines on two axes: (1) proving correct (and only correct) goals, and (2) producing more consistent entailment steps in the tree.

\subsection{Evaluation: Goal Discrimination}

We evaluate our models in two settings, both derived from the validation and test sets of the English-language EntailmentBank dataset. Each setting consists of a 1:1 mixture of examples with valid goals (the original EntailmentBank validation examples) and negative examples with invalid goals, produced by replacing the goal of each positive example with a distinct one drawn from another example. \textbf{Each system is evaluated on whether it can prove the correct goals with valid steps and successfully reject incorrect goals.}

To construct hard negatives, candidate replacement goals are ranked according to TF-IDF weighted overlap between tokens in the destination example and tokens in their original example. For each negative example, the replacement goal with the highest overlap score is selected, excluding goals from examples whose premise sets are subsets of the destination example's premises. We manually check negative examples to ensure they cannot be derived from the provided premises.

In \textbf{Task 1}, examples contain only gold premises (between 2 and 15), while in \textbf{Task 2}, each premise set is expanded to 25 premises through the addition of distractors retrieved from the original premise corpora. We set the $maxSteps$ hyperparameter of the \textsc{SCSearch} algorithm to 20. The maximum gold tree step count in EntailmentBank is 17, so our approach can theoretically recover any binarized gold tree given the right heuristic scores.

Note that we focus our evaluation on this goal discrimination task and validating that individual steps of entailment trees are correct, \emph{not} on recovering the exact trees in EntailmentBank. Our deduction model frequently constructs correct entailment trees that do not match the reference, particularly since our approach is not trained end-to-end on this dataset.

\subsection{End-to-end Baseline}
\label{sec:e2e_desc}

We compare against an \textbf{End-to-end T5} model that we train following the EntailmentWriter paradigm \cite{dalvi-etal-2021-explaining}. The EntailmentWriter system involves fine-tuning a sequence-to-sequence language model to generate an entire entailment tree in linearized form, conditioned on the concatenation of a set of premises and a hypothesis. The EntailmentWriter tree linearization format is shown in Figure \ref{fig:e2e_format}. In the original work, \citet{dalvi-etal-2021-explaining} fine-tune T5-11b \citep{raffel-etal-2020-t5}; we replicate their training setup using T5-3b instead for parity with our other experiments.

In order to evaluate whether an end-to-end model \textit{intrinsically} distinguishes between valid and invalid entailment tree structures, we use the average output confidence over trees generated by the model trained without negative examples, computed as the mean token log-likelihood $\frac{1}{L} \sum_{i=1}^{L} \log p_\theta(t_i \mid t_1 \dots t_{i-1})$. This is motivated by the hypothesis that a model trained as a density estimator for trees composed of sound steps should assign low likelihood to unsound trees. We fit a linear model to predict the goal validity $\in \{0, 1\}$ based on this quantity. We refer to this discriminative setup as \textbf{End-to-end (Intrinsic)}.

We also train a variant of the end-to-end baseline, \textbf{End-to-end T5 (Classify)}, to explicitly predict whether a given goal is valid by including a flag token \texttt{T} or \texttt{F} at the start of the model's output. We augment the model's training data with an equal number of negative examples by randomly resampling goal hypotheses. We prepend \texttt{T} to the target sequence of positive examples, while the target sequence for negative examples is \texttt{F}. Note that this model can predict \texttt{T} and then output a nonsensical entailment tree, as the trees are post-hoc explanations of the decision.

\begin{table*}[h!]
    \centering
    \small
    \begin{tabular}{r c c c c c c}
        \toprule
        & \multicolumn{3}{c}{\textbf{Task 1}} & \multicolumn{3}{c}{\textbf{Task 2}} \\
        \textbf{System} & \textbf{Goal\%} & \textbf{AUROC} & \textbf{\#Steps} & \textbf{Goal\%} & \textbf{AUROC} & \textbf{\#Steps} \\
        \midrule
        Breadth-first          & $33.5\pm0.9$          & $0.88\pm0.0\phantom{0}$ & $3.2\pm2.6$ & $\phantom{0}5.3\pm0.3$ & $0.68\pm0.01$ & $10.6\pm6.1$ \\
        Overlap                & $19.4\pm0.5$          & $0.78\pm0.01$           & $1.8\pm1.2$ & $\phantom{0}1.8\pm0.2$ & $0.61\pm0.02$ & $4.8\pm5.0$ \\
        Overlap (Goal)         & $31.7\pm1.0$          & $0.83\pm0.01$           & $2.4\pm1.6$ & $26.5\pm1.2$           & $0.74\pm0.01$ & $3.5\pm2.8$ \\
        Repetition             & $19.7\pm1.2$          & $0.81\pm0.0\phantom{0}$ & $2.0\pm1.4$ & $\phantom{0}0.9\pm0.4$ & $0.61\pm0.01$ & $2.0\pm1.4$ \\
        Learned                & $28.2\pm1.7$          & $0.83\pm0.01$           & $2.8\pm2.3$ & $\phantom{0}7.9\pm2.0$ & $0.71\pm0.01$ & $6.8\pm4.0$ \\
        Learned (Goal)         & $45.0\pm1.8$          & $0.90\pm0.01$           & $2.7\pm2.1$ & $47.4\pm1.8$           & $0.85\pm0.01$ & $4.5\pm4.1$ \\
        SCSearch     & $48.8\pm1.5$          & $0.91\pm0.01$           & $3.0\pm2.1$ & $52.9\pm1.6$           & $0.85\pm0.01$ & $3.8\pm3.4$ \\ \midrule
        End-to-end (Classify)  & $100.0\pm0.0$         & $0.97\pm0.00$           & $2.8\pm1.6$ & $100.0\pm0.0$          & $0.95\pm0.00$ & $2.2\pm1.2$ \\
        End-to-end (Intrinsic) & -                     & $0.57\pm0.02$           & -           & -                      & $0.62\pm0.02$ & -           \\
        \bottomrule
    \end{tabular}
    \caption{Results from our main experiments on the EntailmentBank test sets. Mean $\pm$ standard deviation is reported for each metric, taken across 10 trials varying the random seed used for nucleus sampling. \textbf{Goal\%} indicates the proportion of valid goals reached by a system's generated trees using the $\alpha=0.81$ threshold. \textbf{AUROC} indicates the area under the receiver operating characteristic when attempting to distinguish gold goals from invalid goals. \textbf{\#Steps} indicates the average number of steps expanded before reaching the goal among trees which reached valid goals; this metric's standard deviation is computed at the example level. See Section \ref{sec:metrics} for more details. }
    \label{tab:main_metrics}
\end{table*}

\begin{table}[h!]
    \centering
    \begin{tabular}{c c}
        \toprule
        \textbf{System} & \textbf{Step Validity} \\
        \midrule
        Learned (Goal) & $74.0\%$ \\
        SCSearch & $\mathbf{82.3\%}$ \\
        End-to-end & $65.0\%$ \\
        \bottomrule
    \end{tabular}
    \caption{Results from manual annotation of step validity. Steps are sampled from inference on Task 2. Macro-average inter-annotator agreement (Cohen's $\kappa$) is $0.72$.}
    \label{tab:manual_eval}
\end{table}

\begin{figure*}[h!]
    \centering
    \includegraphics[width=\columnwidth]{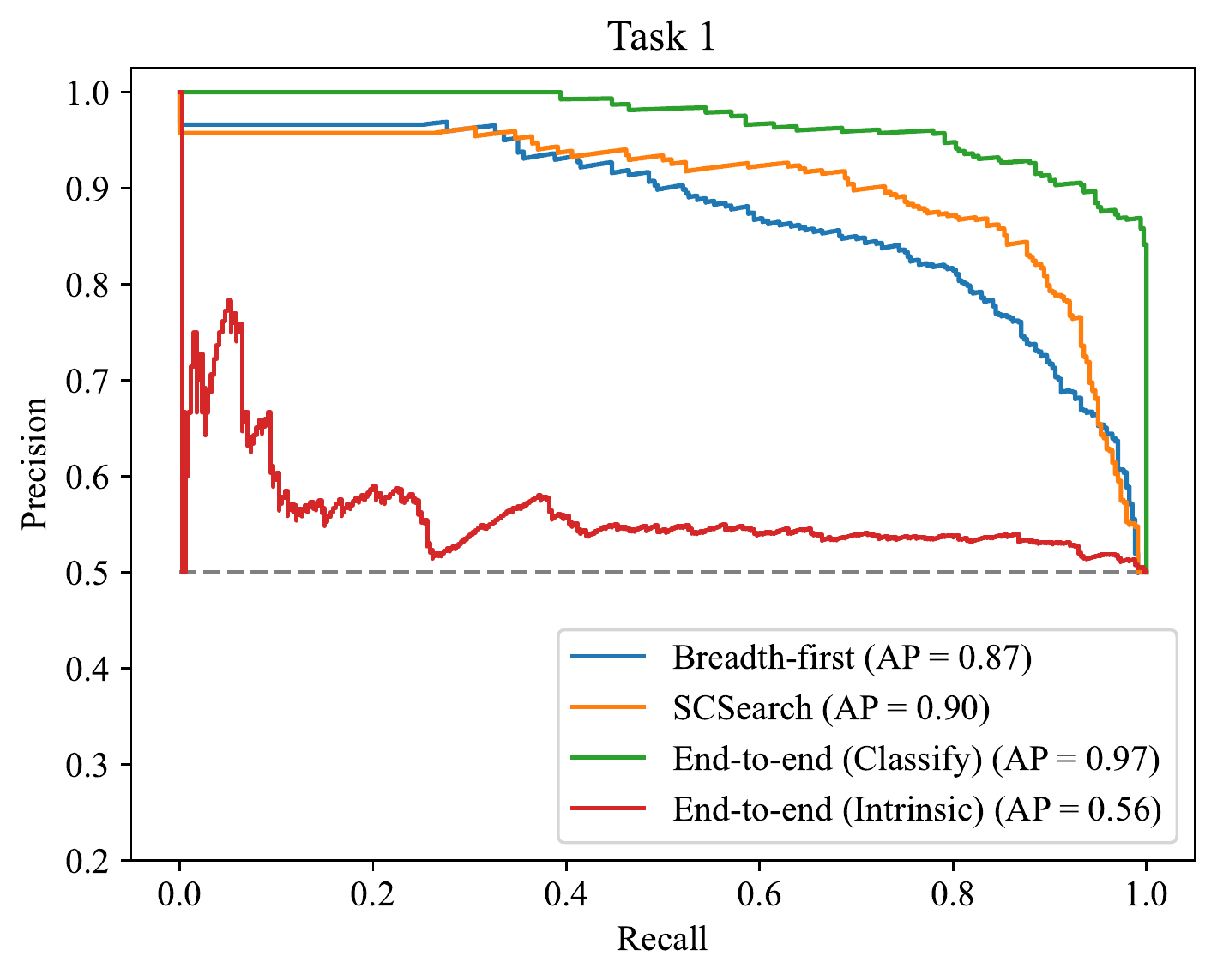}
    \includegraphics[width=\columnwidth]{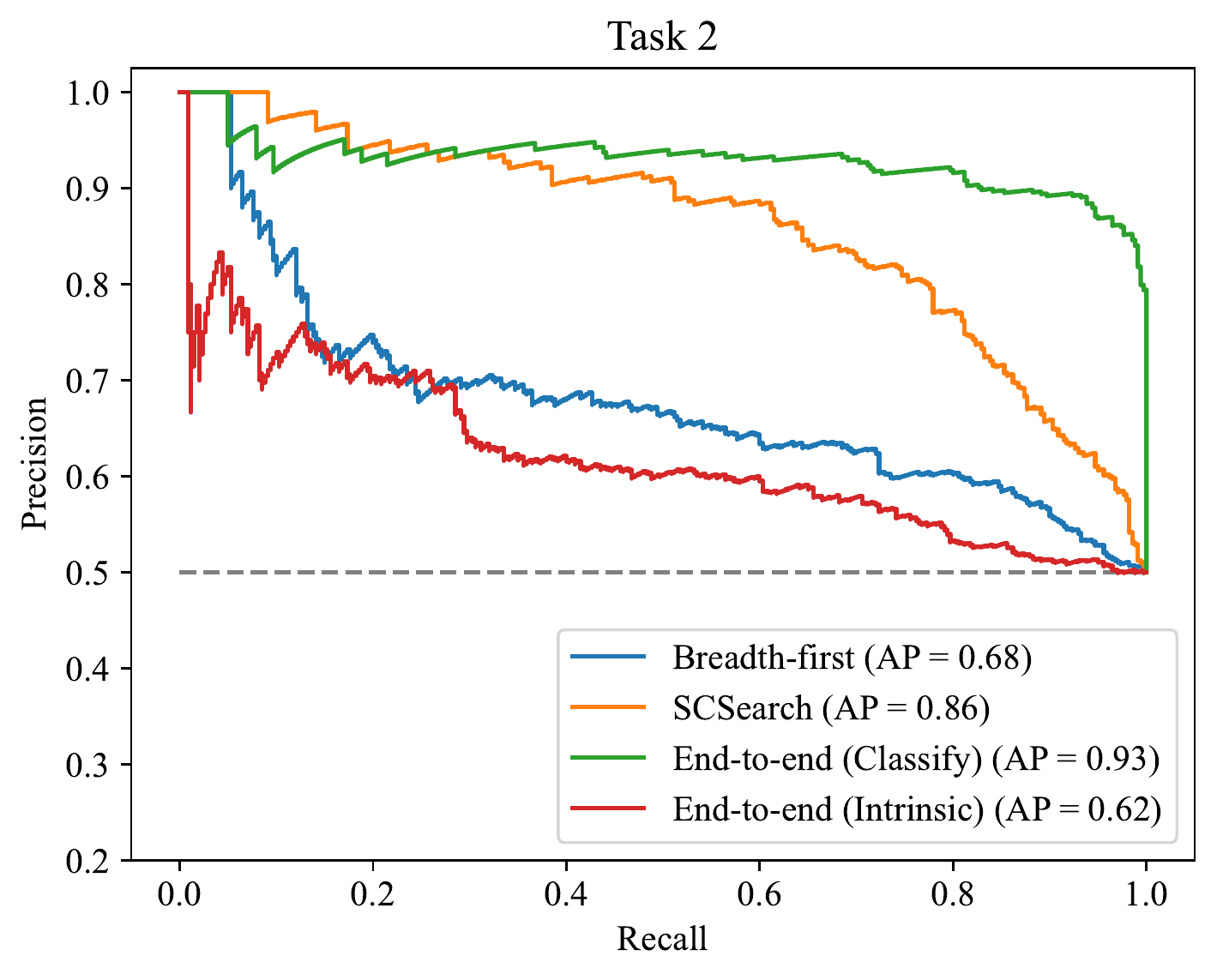}
    \caption{Precision-recall curves for breadth-first search, our full method, and the end-to-end baseline. The dashed line at precision$=0.5$ corresponds to random chance.}
    \label{fig:curves}
\end{figure*}

\subsection{Metrics}
\label{sec:metrics}

For both Task 1 and Task 2, consistent models should be able to reach gold goals while also failing to prove invalid goals. To measure the former, we report the number of valid goals reached by each system as \textbf{Goal\%}. For our search systems, this metric is computed as the proportion of positive examples for which any generated conclusion had a goal entailment score higher than the threshold $\alpha=0.81$ (see Section \ref{sec:impl_training}).
Goal\% scores for the end-to-end model correspond to its self-reported success rate --- the proportion of positive examples for which the model emits the \texttt{T} token.

We report the average number of steps expanded before reaching a valid goal as \textbf{\#Steps}. This metric is averaged over examples for which a system is able to reach the goal.

To measure whether systems are able to distinguish invalid goals from valid goals, we also compute precision-recall curves and report the area under the receiver operating characteristic (\textbf{AUROC}) of each system for both tasks. We produce these curves for search models by varying the goal entailment threshold $\alpha$. For the \textbf{End-to-end (Intrinsic)} model, we vary a threshold on the average generated token likelihood, and for the \textbf{End-to-end (Classify)} model, we vary a threshold on the value of $p(\texttt{T}) / (p(\texttt{T})+p(\texttt{F}))$, the score assigned to the ``valid'' flag token by the model out the two possible validity flags.

In addition to evaluating end-to-end performance through the above tasks, we would also like to understand the internal consistency of generated trees. To that end, we conduct a manual study of step validity. We sample 100 steps uniformly across valid-goal Task 2 examples for each of three systems: our full system, our system without mid-training, and the end-to-end system. The resulting set of 300 steps is shuffled and triply labeled by three annotators without knowledge of which examples came from which system. For each example, the annotators assess whether a single step's conclusion can be inferred from the input premises with minimal additional world knowledge. We discuss these results in Section~\ref{sec:step_validity}.

\subsection{Implementation and Training}
\label{sec:impl_training}

All sequence-to-sequence language models we experiment with are instances of T5-3b \cite{raffel-etal-2020-t5} with 3 billion parameters. Bridging entailment models and learned heuristic models are derived from DeBERTa Large \cite{he2021deberta} with 350M parameters. Fine-tuning is performed with the Hugging Face \texttt{transformers} library \cite{wolf-etal-2020-transformers}. Further details including fine-tuning hyperparameters are included in the appendix.

\paragraph{Step deduction model} We train our step deduction model using gold steps sampled from trees in the EntailmentBank (\textbf{EB}) training split, totaling 2,762 examples. Our full system, which we refer to as \textbf{SCSearch}, is also ``mid-trained'' on ParaPattern substitution data \cite{bostrom-etal-2021-flexible}. The ParaPattern data is derived semi-synthetically from English Wikipedia, totaling $\sim$120k examples. In the mid-training configuration, an instance of T5 is fine-tuned for one epoch on the ParaPattern data and then for one epoch on the EntailmentBank data, after which optimal validation loss is reached.

\paragraph{Learned heuristic models}

Data for our learned heuristic models is constructed by taking step inputs from the EntailmentBank training set's gold trees. For each positive step example, a corresponding negative example is produced by replacing one input statement with a sentence drawn at random from all statements in the training set that do not appear in the original step's subtree. Examples used to train the Learned (Goal) heuristic additionally contain the original step's gold goal concatenated to their input. Our full SCSearch system uses the Learned (Goal) heuristic.

\section{Results}

The results of our experiments on Task 1 and Task 2 are shown in Table \ref{tab:main_metrics}. Table \ref{tab:manual_eval} shows the results of our manual step validity evaluation.

\subsection{Complete Deductions}

\paragraph{The end-to-end approach is not a sound model of entailment tree structure.}
End-to-end T5 (Classify) is nominally able to ``prove'' 100\% of gold goals by finishing a proof with \texttt{-> hypot}. However, as shown in Figure \ref{fig:curves}, using the generation confidence of End-to-end T5 to discriminate between valid goals and invalid goals (the Intrinsic method) is not much better than random chance, since the model has similar confidence when ``proving'' invalid goals as it does when generating trees for valid goals. This means that the model's output distribution does not penalize the generation of invalid steps.

\paragraph{Our approach is able to prove most goals while exhibiting much better internal consistency than the end-to-end approach.} Our full SCSearch system nearly matches the performance of the End-to-end (Classify) system on Task 1 and Task 2, losing chiefly in high-recall settings. Critically, Section~\ref{sec:step_validity} will show that it achieves a much higher rate of step validity in the process.

Mid-training the step deduction model also increases the proportion of reachable valid goals by 3-5\% (compare Learned (Goal) to SCSearch in Table~\ref{tab:main_metrics}). It is worth noting that the chosen threshold for our goal entailment module sacrifices recall in favor of avoiding false positives, as shown in Table~\ref{tab:goalbridge}, meaning that our reported Goal\% rate is an underestimate. In Figure~\ref{fig:curves}, we see that our SCSearch method can achieve 80\% recall at roughly 80\% precision with a slightly lower threshold.

\paragraph{A goal-oriented heuristic is critical.}
If we compare Breadth-first, Repetition, our two Overlap methods, and our two Learned heuristics, both Table~\ref{tab:main_metrics} and Figure~\ref{fig:curves} show that the incorporating goal information into the planning process is essential for Task 2, as only the Overlap (Goal) and Learned (Goal) heuristics are able to reach a reasonable number of valid goals in the presence of distractor premises. We can see in Figure~\ref{fig:curves} how much breadth-first degrades from Task 1 to Task 2, largely due to timing out.

\subsection{Individual Step Validity}
\label{sec:step_validity}

The most crucial divergence between our method and the end-to-end method arises in the evaluation of individual steps. The End-to-end (Classify) method can produce correct decisions, and as shown in Figure~\ref{fig:e2e_format} it can always claim to have produced the goal statement, but is its reasoning sound?

Table~\ref{tab:manual_eval} shows that our best SCSearch model produces valid inferences $82\%$ of the time, improving by $17\%$ absolute over the end-to-end model as well as improving over the version of our system without mid-training on ParaPattern. This result confirms our hypothesis about the end-to-end method: despite often predicting the right label and predicting an associated entailment tree, these entailment trees are not as likely to be valid due to their post-hoc nature and the conflation of search and generation.  

We can use our step validity rate to approximate the expected number of fully-valid trees based on the observed depth distribution. Under the conservative assumption that observed errors are distributed uniformly w.r.t. depth, the expected number of fully valid trees in a dataset $D = \{T_1\dots T_{|D|} \}$ for validity rate $v$ can be computed as
$ \frac{1}{|D|} \sum_{i=1}^{|D|} v^{|T_i|}$.
At a step validity rate of 82\% we should expect $\sim$58\% of trees generated by our system to be error-free. Under the same assumption, according to the end-to-end model's step validity rate we should expect only $\sim$35\% of its trees to be fully valid.

In Section~\ref{sec:err_step} we examine observed error patterns in invalid steps. Crucially, even when our system produces a tree involving an invalid step, it is easy to audit the tree and determine exactly where the reasoning error occurred, since each step is conditioned only on its immediate premises. In contrast, the end-to-end model attends to all premises and the hypothesis at every step, meaning that when an inconsistent step is generated, it is difficult to diagnose the cause.

\subsection{Goal Entailment}

\begin{table}[ht]
\renewcommand{\tabcolsep}{1.0mm}
    \centering
    \small
    \begin{tabular}{c c c c c}
        \toprule
        \textbf{Model} & $\alpha$ & \textbf{F1} & \textbf{Precision} & \textbf{Recall}\\
        \midrule
        DeBERTa WANLI & 0.92 & 71.8 & 65.1 & \textbf{80.0} \\
        +\textsc{EBEntail} & 0.83 & 72.4 & 91.3 & 60.0 \\ 
        +\textsc{EBEntail-Active} & 0.81 & \textbf{77.9} & \textbf{95.8} & 65.7 \\
        \bottomrule
    \end{tabular}
    \caption{Evaluation of goal-bridging entailment models on \textsc{EBEntail}. Both of the models fine-tuned on \textsc{EBEntail} are evaluated using 3-fold cross validation. $\alpha$ indicates the best entailment score threshold.}
    \label{tab:goalbridge}
\end{table}

\noindent
The results in Table~\ref{tab:main_metrics} depend on an accurate assessment of when we have successfully deduced the hypothesis. To that end, we evaluate our goal entailment model against labeled test data. Table~\ref{tab:goalbridge} shows the results of our evaluation on the \textsc{EBEntail} dataset described in Section \ref{sec:goal_ent}. We view precision as more important than recall, as a stringent criteria for determining whether a tree has reached the goal increases confidence in our evaluation results.

Our best F1 score, using \textsc{EBEntail-Active}, is only slightly lower than the lowest F1 score of the annotators when evaluated against the majority vote. Annotator agreement is moderate; macro-averaged inter-annotator F1 is 0.83 and Cohen's $\kappa$ is 0.54. This indicates that the problem of determining when a statement straightforwardly entails the goal is subjective; the boundary between `trivial' entailment and a case which needs an additional reasoning step is somewhat fuzzy.

\subsection{Error Analysis: Step Model}
\label{sec:err_step}

Although our step model cannot hallucinate based on the hypothesis, it can still fail to produce valid intermediate steps due to other challenges.

One error type we see is \textbf{indiscriminate unification}: when given incompatible premises, the step model will sometimes still attempt to combine them, resulting in improper conclusions. For example, given the premises \textit{``Earthquake can change the earth's surface. In a short amount of time is similar to rapidly.''} one conclusion generated by the model is \textit{``Earthquake changes the earth's surface rapidly.''} This could be avoided through the use of more selective heuristics, or by explicitly supervising step models with negative examples in order to encourage conservative conclusions in these cases.

We also observe \textbf{compounding errors} in conclusions generated from erroneous premises. For example, given the premises \textit{``Offspring will inherit a scar on both knees except not both knees. Offspring will inherit a scar on the knee from parents.''} the model generates \textit{``Parents will inherit a scar on both knees.''} This kind of relation assignment mistake is uncommon outside of instances involving bad premises. These errors could potentially be mitigated by training heuristics to avoid corrupted premises.

\subsection{Error Analysis: Goal Entailment}

An additional source of error arises from cases where the goal entailment model is unable to predict the correct entailment relationship between an output from the step model and the goal hypothesis.

One reason this arises is \textbf{definitional knowledge}. When given the premise \textit{``Sugar is soluble in water,''} the model does not predict entailment of the goal hypothesis \textit{``Sugar cubes will dissolve in water when they are combined.''} English speakers who know the definition of `soluble' and recognize that sugar cubes are made of sugar could reasonably understand this as entailed. However, the degree of definitional knowledge that should be expected of the entailment model is subjective and often a source of annotator disagreement.

Another cause of errors are \textbf{ill-formed statements}. For example, the model predicts that \textit{``Lichens and soil are similar to being produced by breaking down rocks.''} entails \textit{``Lichens breaking down rocks can form soil.''} However, it is unclear what is ``similar'' in the generated statement due to poor syntax. Labels for examples like this often vary depending on how the annotator understood the step model's output. Improving the step model to reduce compounding generation errors will mitigate this issue.

Finally, the entailment may sometimes be predicated on \textbf{context}. The model predicts that \textit{``A new moon will occur on june 30 when the moon orbits the earth.''} entails \textit{``The next new moon will occur on june 30.''} In this case, the model is assuming that `a new moon' occurring is equivalent to `the next new moon' occurring. Depending on annotator assumptions, cases like this can also be somewhat subjective.

Future work could expand the training data of our NLI model to account for the subjectivity of NLI judgments \cite{pavlick-kwiatkowski-2019-inherent,chen-etal-2020-uncertain,nie-etal-2020-learn}, particularly by modifying our data collection procedure \cite{zhang-etal-2021-learning-different}.

\section{Related work}

Our work reflects an outgrowth of several lines of work in reading comprehension and textual reasoning.
Multi-hop question answering models (\citealt{chen2021multihop, min-etal-2019-multi, nishida-etal-2019-answering}, inter alia) also build derivations linking multiple statements to support a conclusion. However, these models organize selected premises into a chain or leave them unstructured as opposed to composing them into an explicit tree. 

The NLProlog system \cite{weber-etal-2019-nlprolog} frames multi-hop reading comprehension explicitly as a proof process, performing proof search using soft rule unification over vector representations of predicates and arguments. Similar backward search ideas were used in \citet{arabshahi2021conversational}. PRover \cite{saha-etal-2020-prover} and ProofWriter \cite{tafjord-etal-2021-proofwriter} also frame natural language deduction as proof search, although both systems are evaluated in a synthetic domain of limited complexity. \citet{Betz2021DeepA2AM} also use synthetic data to improve reasoning models through mid-training, although the improvements they observe are limited to premise selection performance.

\citet{hu-etal-2020-monalog} and \citet{chen-etal-2021-neurallog} propose systems which perform single-sentence natural language inference through proof search in the natural logic space. Our work also relates to earlier efforts on natural logic \cite{MacCartneyManning2009,angeli-etal-2016-combining} but is able to cover far more phenomena by relaxing the strict constraints of this framework. Finally, the Leap of Thought system \cite{talmor-leap-of-thought} tackles some related ideas in a discriminative reasoning framework.

The recent chain-of-thought \cite{wei2022-cot} and Scratchpads \cite{nye21-scratchpads} methods also generate intermediate text as part of answer prediction. However, like the end-to-end baseline we consider, these techniques are free to generate unsound derivations. Published results with these techniques are strongest for tasks involving mathematical reasoning or programmatic execution, whereas on textual reasoning datasets like StrategyQA \cite{geva-etal-2021-aristotle} they only mildly outperform a few-shot baseline.

\section{Discussion and Conclusion}

In this work, we propose a system that performs natural language reasoning through generative deduction and heuristic-guided search. We demonstrate that our system produces entailment trees that are more internally consistent than those of an end-to-end model, and that its factored design allows it to successfully prove valid goals while being unable to hallucinate trees for invalid goals. We believe that this modular deduction framework can be readily extended to empower future reasoning systems.

\section{Limitations}

The baseline approach we consider in this work, end-to-end modeling of entailment tree generation, enjoys the convenience of simple inference and quadratic complexity. However, the computational overhead of sequence-to-sequence models places a hard limit on the tree size and premise count that can be handled in the end-to-end setting; moreover, recent results call into question how well end-to-end Transformers can generalize this type of reasoning \cite{zhang22paradox}. Our structured approach allows arbitrarily large premise sets and step counts. However, by discretizing the reasoning in the SCSearch procedure, we do face a runtime theoretically exponential in proof size to do exhaustive search. In practice, we limit our search to a finite horizon and find that this suffices to provide a practical wall clock runtime, never exceeding 5 seconds for any single example. Future work on higher tree depths may have to reckon with the theoretical limitations of this procedure, possibly through the use of better heuristics.

Our experiments are conducted exclusively on English datasets. While we hypothesize that our approach would work equally well for another language given a pretrained sequence-to-sequence model for that language with equivalent capacity, such models are not available universally across languages, representing an obstacle for transferring our results to languages beyond English.

Furthermore, the EntailmentBank dataset on which we train and evaluate targets the elementary science domain, raising a question of domain specificity. In future work, we plan to evaluate deduction models on additional datasets with different style, conceptual content, and types of reasoning in order to verify that the factored approach is equally applicable across diverse settings.

\section*{Acknowledgments}

This work was supported by NSF CAREER Award IIS-2145280, a grant from Open Philanthropy, and gifts from Salesforce and Adobe. This material is also based on research that is in part supported by the Air Force Research Laboratory (AFRL), DARPA, for the KAIROS program under agreement number FA8750-19-2-1003. The views and conclusions contained herein are those of the authors and should not be interpreted as necessarily representing the official policies, either expressed or implied, of DARPA, or the U.S. Government. The U.S. Government is authorized to reproduce and distribute reprints for governmental purposes notwithstanding any copyright annotation therein. Thanks to the anonymous reviewers for their helpful feedback.

\bibliography{custom}
\bibliographystyle{acl_natbib}
\newpage
\appendix

\setlength{\parindent}{0pt}

\section{Implementation Details}
\label{sec:appendix}

Hugging Face \texttt{transformers} version: 4.19.1

Two NVidia Quadro 8000 GPUs were used for all experiments in this paper.

\begin{table}[h!]
\small
\begin{tabular}{r c}
    \toprule
    \textbf{Hyperparameter} & \textbf{Value} \\
    \midrule
    Base model & \href{https://huggingface.co/t5-3b}{T5-3b} \\
    Total batch size & 8 \\
    Initial LR & 5e-5 \\
    Epoch count & 3 (early stopping on val. loss) \\
    \bottomrule
\end{tabular}
\caption{End-to-end model fine-tuning configuration (hyperparameters left at \texttt{transformers} default if unspecified)}
\end{table}

\begin{table}[h!]
\small
\begin{tabular}{r c}
    \toprule
    \textbf{Hyperparameter} & \textbf{Value} \\
    \midrule
    Base model & \href{https://huggingface.co/t5-3b}{T5-3b} \\
    Total batch size & 12 \\
    Initial LR & 5e-5 \\
    Epoch count & 2 (early stopping on val. loss) \\
    \bottomrule
\end{tabular}
\caption{EntailmentBank-only step deduction model fine-tuning}
\end{table}

\begin{table}[h!]
\small
\begin{tabular}{r c}
    \toprule
    \textbf{Hyperparameter} & \textbf{Value} \\
    \midrule
    Base model & \href{https://huggingface.co/t5-3b}{T5-3b} \\
    Total batch size & 12 \\
    Initial LR & 5e-5 \\
    Epoch count & 1 \\
    \bottomrule
\end{tabular}
\caption{Step deduction model mid-tuning on ParaPattern-substitution}
\end{table}

\begin{table}[h!]
\small
\begin{tabular}{r c}
    \toprule
    \textbf{Hyperparameter} & \textbf{Value} \\
    \midrule
    Base model & \href{https://huggingface.co/microsoft/deberta-large-mnli}{DeBERTa Large MNLI} \\
    Total batch size & 32 \\
    Initial LR & 1e-5 \\
    Epoch count & 1 \\
    \bottomrule
\end{tabular}
\caption{Goal entailment model fine-tuning on \textsc{EBEntail}+\textsc{EBEntail-Active}}
\end{table}

\begin{table}[h!]
\small
\adjustbox{width=\columnwidth}{
\begin{tabular}{r c}
    \toprule
    \textbf{Hyperparameter} & \textbf{Value} \\
    \midrule
    Base model & \href{https://huggingface.co/microsoft/deberta-v3-large}{DeBERTa-v3 Large} \\
    Total batch size & 32 \\
    Initial LR & 2e-5 \\
    Epoch count & \begin{tabular}{@{}c@{}}2 (no goals)\\7 (w/goals)\end{tabular} (early stopping on val. loss) \\
    \bottomrule
\end{tabular}}
\caption{Learned heuristic model fine-tuning}
\end{table}

\newpage

\onecolumn
\section{Examples}
\label{sec:app_exs}

\begin{tabular}{@{}c@{}}
\includegraphics[width=\textwidth]{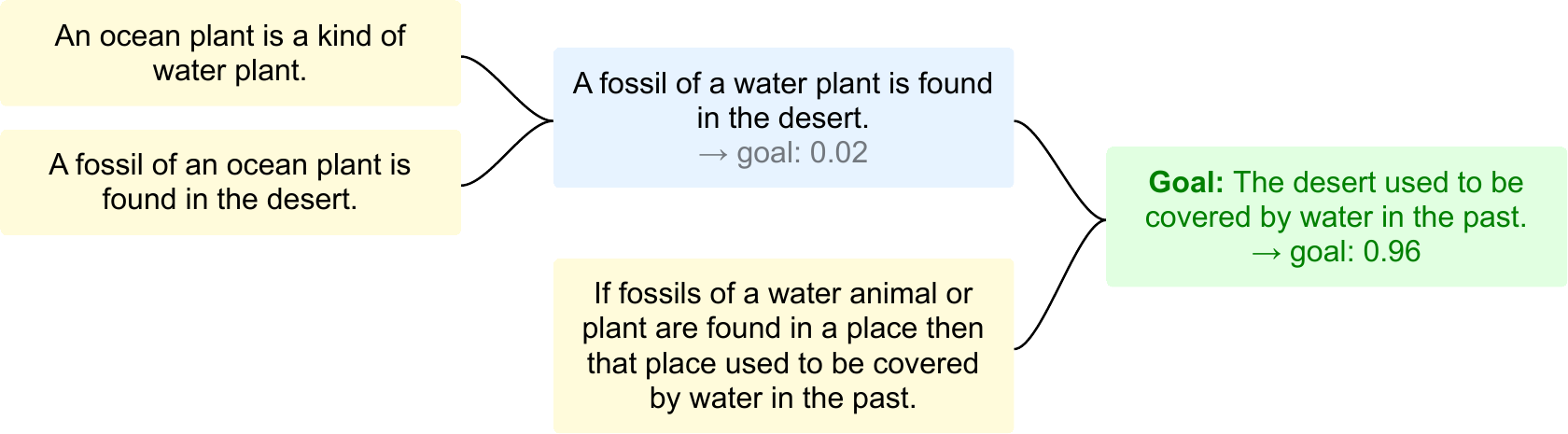} \\
\midrule
\includegraphics[width=\textwidth]{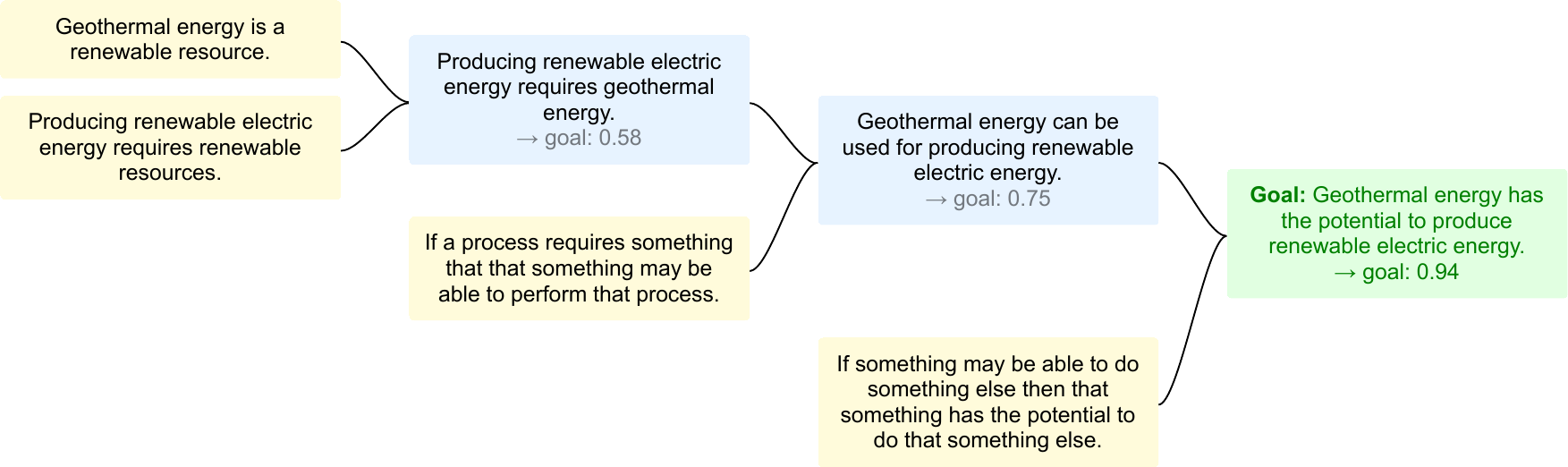} \\
\midrule
\includegraphics[width=\textwidth]{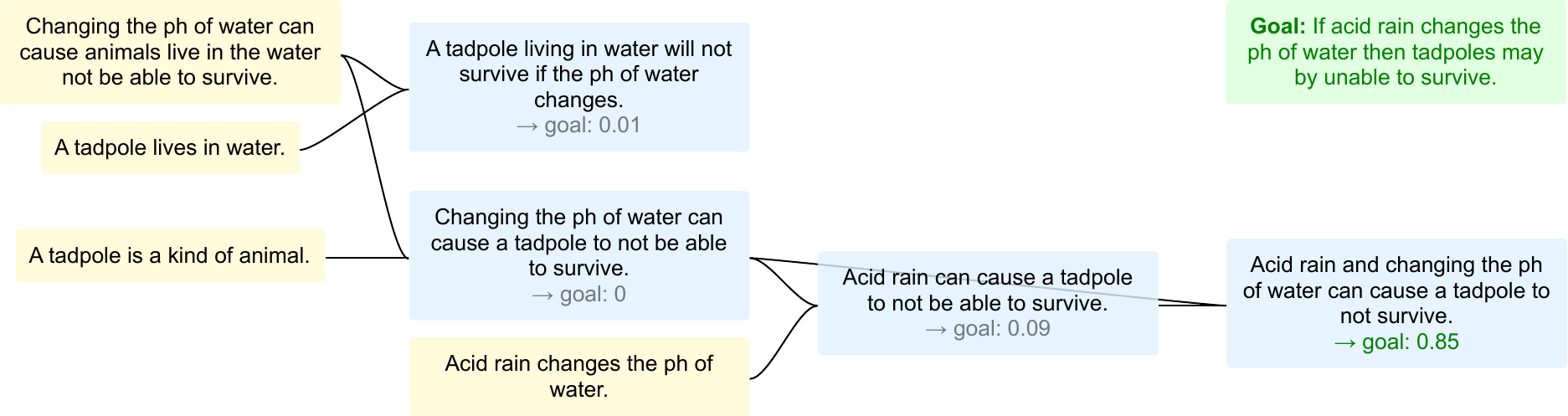}

\end{tabular}
\captionof{figure}{Examples of entailment trees generated by our SCSearch system. In the second example, the first intermediate conclusion exhibits an overzealous substitution, unifying `renewable resources' in `[...] requires renewable resources' with `geothermal energy' despite geothermal energy being just one example of a renewable resource. This is one form of the indiscriminate unification issue noted in Section \ref{sec:err_step}. The last example includes an additional step sampled outside of the successful entailment tree; this step technically entails the goal but goes undetected by the goal entailment module.}

\end{document}